\documentclass[10pt,twocolumn,letterpaper]{article}

\usepackage[pagenumbers]{cvpr} 

\usepackage{graphicx}
\usepackage{amsmath}
\usepackage{amssymb}
\usepackage{booktabs}

\usepackage[pagebackref,breaklinks,colorlinks]{hyperref}

\usepackage[capitalize]{cleveref}
\crefname{section}{Sec.}{Secs.}
\Crefname{section}{Section}{Sections}
\Crefname{table}{Table}{Tables}
\crefname{table}{Tab.}{Tabs.}

\def\cvprPaperID{9210} 
\def\confName{CVPR}
\def\confYear{2023}

\begin{document}

\title{Can lies be faked? Comparing low-stakes and high-stakes deception video datasets from a Machine Learning perspective}

\author{Mateus Karvat Camara\\
Universidade Estadual do Oeste do Paraná\\
{\tt\small mkarvat@gmail.com}
\and
Adriana Postal\\
Universidade Estadual do Oeste do Paraná\\
{\tt\small adriana.postal@unioeste.br}
\and
\hspace{1.2cm}Tomas Henrique Maul             \\
\hspace{1.2cm}University of Nottingham           \\
\hspace{1.2cm}{\tt\small tomas.maul@nottingham.edu.my}
\and
\hspace{0.8cm}Gustavo Paetzold\\
\hspace{0.8cm}Universidade Tecnológica Federal do Paraná\\
\hspace{0.8cm}{\tt\small ghpaetzold@utfpr.edu.br}
}
\maketitle

\begin{abstract}
Despite the great impact of lies in human societies and a meager 54\% human accuracy for Deception Detection (DD), Machine Learning systems that perform automated DD are still not viable for proper application in real-life settings due to data scarcity. Few publicly available DD datasets exist and the creation of new datasets is hindered by the conceptual distinction between low-stakes and high-stakes lies. Theoretically, the two kinds of lies are so distinct that a dataset of one kind could not be used for applications for the other kind. Even though it is easier to acquire data on low-stakes deception since it can be simulated (faked) in controlled settings, these lies do not hold the same significance or depth as genuine high-stakes lies, which are much harder to obtain and hold the practical interest of automated DD systems. To investigate whether this distinction holds true from a practical perspective, we design several experiments comparing a high-stakes DD dataset and a low-stakes DD dataset evaluating their results on a Deep Learning classifier working exclusively from video data. In our experiments, a network trained in low-stakes lies had better accuracy classifying high-stakes deception than low-stakes, although using low-stakes lies as an augmentation strategy for the high-stakes dataset decreased its accuracy.
\end{abstract}

\section{Introduction}
\label{sec_intro}

Lies are widespread throughout human societies and it is estimated that humans lie more than twice a day \cite{DePaulo1996Lying}. In contexts such as Justice, lies can have meaningful consequences, with 53\% of the exonerations in the United States between 1989 and 2012 involving deception \cite{Gross2012Exonerations}. Therefore, researchers have been interested in Deception Detection (DD) for centuries \cite{Khan2021Deception}, and, since human accuracy for this task is merely 54\% \cite{bond2006}, attempts have been made to build automated systems for DD, the most successful being the polygraph. However, despite its widespread use in courts, the polygraph has proven to be an unreliable system \cite{Fiedler2002WhatIs,Honts1994Mental} and its usage in official settings is currently not recommended.

The quest for automated DD systems has been reinvigorated by the success of Machine Learning (ML) applications and many researchers have directed efforts toward this field. Even though some papers achieved impressive accuracies above the 90\% mark \cite{Venkatesh2020Video, Karnati2022Lienet, Ding2018Face, Krishnamurthy2018DL, Gogate2017DL, Carissimi2018MultiView, Chebbi2021Deception, Wu2017DD}, the proper application of ML systems to DD in real-life is still not viable due to data scarcity \cite{Mambreyan2022Dataset,Khan2021Deception}. Few video-based DD datasets have been created and even fewer made available for public use, hindering the development of proper techniques and experiments in the area. But most importantly, the existing datasets lack sufficient data for use in real-life settings, with no public DD dataset having more than 400 videos, which is far from enough to properly train a system capable of correctly identifying lies outside of experimental settings \cite{Khan2021Deception}.

Aside from the inherent challenges of building datasets based on human subjects such as ethics and biases, DD has a conceptual barrier that complicates matters further: the distinction between low-stakes and high-stakes lies \cite{vrij2008detecting}. According to the literature \cite{Hartwig2014Lie}, there is a significant distinction between lies told in different contexts since cues for deception will emerge in high-stakes situations in which the liars develop a stronger emotional response, a phenomenon known as non-verbal leakage. According to Frank and Ekman \cite{Frank1997Ability}, ``it is the presence of these emotions, such as guilt, fear of being caught, and disgust, that can betray the liar’s deception when they are leaked through nonverbal behaviors such as facial expressions [...] or voice tone''. In this sense, trials, interviews or negotiations could be considered as high-stakes scenarios, while contexts that do not elicit such strong reactions could be seen as low-stakes: truth games, roleplaying or even situations in which participants are instructed to lie by researchers.

Even though low-stakes lies are more frequent in human lives and can be easily simulated (faked) in controlled settings, the practical interest for automated DD falls on high-stakes lies due to their potential impact. Based on that, solving the issue of data scarcity for DD would mean creating a large and diverse public dataset of high-stakes lies. However, the acquisition of this type of data poses ethical and privacy issues and, most importantly, given that human accuracy for DD is so low, any labeling of these samples should be based on unassailable proof that testimonies were either truthful or deceptive.

Ideally, ML systems would be trained on fake lies easily acquired in controlled settings (low-stakes lies) and applied to high-stakes DD scenarios, though the conceptual distinction between these two kinds of lies has blocked such possibility. We then arrive at an impasse since, on one hand, high-stakes datasets are needed for real-life applications but prove difficult to be built, and, on the other hand, low-stakes datasets can be created more easily from lies told in fake scenarios yet have limited practical relevance.

Despite being a possible solution to such an impasse, to the best of our knowledge, no study has evaluated if there is, indeed, a significant distinction between low-stakes and high-stakes deception from an ML perspective, from which two questions of practical significance arise: ``Can an ML system trained in a low-stakes deception dataset be used to classify high-stakes lies?'' and ``Can low-stakes lies be used for data augmentation in high-stakes deception datasets?''.

Hence, this paper investigates this distinction by comparing the performance of a Deep Learning system with two datasets: Real-life Trial (RLT) \cite{PerezRosas2015DD}, a high-stakes dataset with footage from trials and currently considered the standard video-based DD dataset \cite{Mathur2020Introducing}; and Box of Lies (BoL) \cite{Soldner2019Box}, a low-stakes dataset with videos taken from a truth game on the television show ``The Tonight Show Starring Jimmy Fallon \textregistered''. Focusing exclusively on video data from each dataset and performing binary classification, a network based on the Slowfast architecture \cite{Feichtenhofer_2019_ICCV} is created and several experiments are performed to evaluate if the aforementioned distinction does indeed hold true based on experimental results. We hope that, by examining this distinction, researchers may redirect their efforts toward building large-scale deception datasets while making the best use of the limited data available in this important application area.

Thus, this paper is organized as follows: \cref{sec_datasets} discusses the existing public video-based DD datasets and ML papers that use them, \cref{sec_methodology} describes our methodology, \cref{sec_results} presents and discusses the results from our experiments and \cref{sec_conclusion} summarizes our findings.

\section{Deception detection datasets}
\label{sec_datasets}

Back in 2012, Gokhman \etal~\cite{Gokhman2012Search} and Fitzpatrick and Bachenko \cite{Fitzpatrick2012Building} discussed the development of a standard Deception Detection (DD) dataset, since none was available at the time, which required every paper to build its own dataset, limiting comparisons between studies and hindering the advancement of the field. Since then, many papers have been published with experiments conducted in video-based datasets which were not publicly available \cite{Monaro2022Detecting, Su2016Does, Belavadi2020Multimodal, Khan2021Deception, Bhaskaran2011Lie} due to the sensitive nature of the task. However, four video-based public datasets (presented in \cref{tab_datasets}) have been created and are discussed here.

We focus on datasets with videos for their simplicity of being deployed in real-life situations, in contrast to Electroencephalogram (EEG) or Functional near-infrared spectroscopy (fNIRS) which require special equipment to acquire data, and for the fact that they present the best results, in relation to audio and text, in papers that compare the performance of unimodal classification \cite{Krishnamurthy2018DL, Mathur2020Introducing, Wu2017DD}. Yet, several papers have worked with DD datasets from other modalities (some of which are publicly available), such as EEG \cite{Baghel2020Truth, Amber2019P300, Dodia2020Lie}, fNIRS \cite{HernandezReynoso2013DD}, audio \cite{Xue2019, Mendels2017Hybrid, GonzaleBillandon2019CanRobot} and text \cite{Ruiter2018Mafiascum, HernandezCastaneda2017CrossDomain, Delgado2021DD, Ho2019Context}.

\begin{table*}
\centering
\begin{tabular}{ccccc}
\hline
\textbf{Dataset}                                                                       & \textbf{Category} & \textbf{Videos} & \textbf{Individuals} & \textbf{Observations}                                                              \\ \hline
Real-life Trial \cite{PerezRosas2015DD}                                                                       & High-stakes       & 121 (110)            & 56 (51)                  & \begin{tabular}[c]{@{}c@{}}Standard dataset for\\ Deception Detection \cite{Mathur2020Introducing}\end{tabular} \\
Box of Lies \cite{Soldner2019Box}                                                                           & Low-stakes        & 68 (93)             & 26 (33)                  & -                                                                                  \\
\begin{tabular}[c]{@{}c@{}}Miami University Deception\\Detection Dataset \cite{Lloyd2019Miami}\end{tabular} & Low-stakes        & 320             & 80        & Balanced dataset              \\
Bag-of-Lies    \cite{Gupta2019Bagoflies}                                                                        & Low-stakes        & 325             & 35                   & \begin{tabular}[c]{@{}c@{}}Includes EEG and\\ Gaze information\end{tabular}        \\ \hline
\end{tabular}
\caption{Publicly available video datasets for Deception Detection. Numbers in parentheses are presented for reference and correspond to the numbers used in this paper for the respective datasets.}
\label{tab_datasets}
\end{table*}

\subsection{Low-stakes datasets}
\label{subsec_low}

To the best of our knowledge, there are currently three publicly available video-based low-stakes DD datasets: Bag-of-Lies \cite{Gupta2019Bagoflies}, Miami University Deception Detection Dataset (MU3D) \cite{Lloyd2019Miami} and Box of Lies (BoL) \cite{Soldner2019Box}.

Bag-of-Lies \cite{Gupta2019Bagoflies} is described by its authors as a ``casual deception'' dataset, containing 325 videos of 35 volunteers that had to describe images from a selected set being free to be truthful or deceptive. Being a multimodal dataset, it has visual, audio and gaze data (acquired with an eye tracker) for all testimonies, and EEG data for 201 videos.

The MU3D \cite{Lloyd2019Miami} was created with the goal of being a standardized, unbiased and balanced dataset, being made of 320 videos from 80 subjects (20 Black female, 20 Black male, 20 White female, and 20 White male), each one having 2 lies and 2 truths. In each video, participants describe people they like and people they dislike, with all videos having audio and accompanying transcriptions.

\begin{figure}
   \centering
   \includegraphics[width=\linewidth]{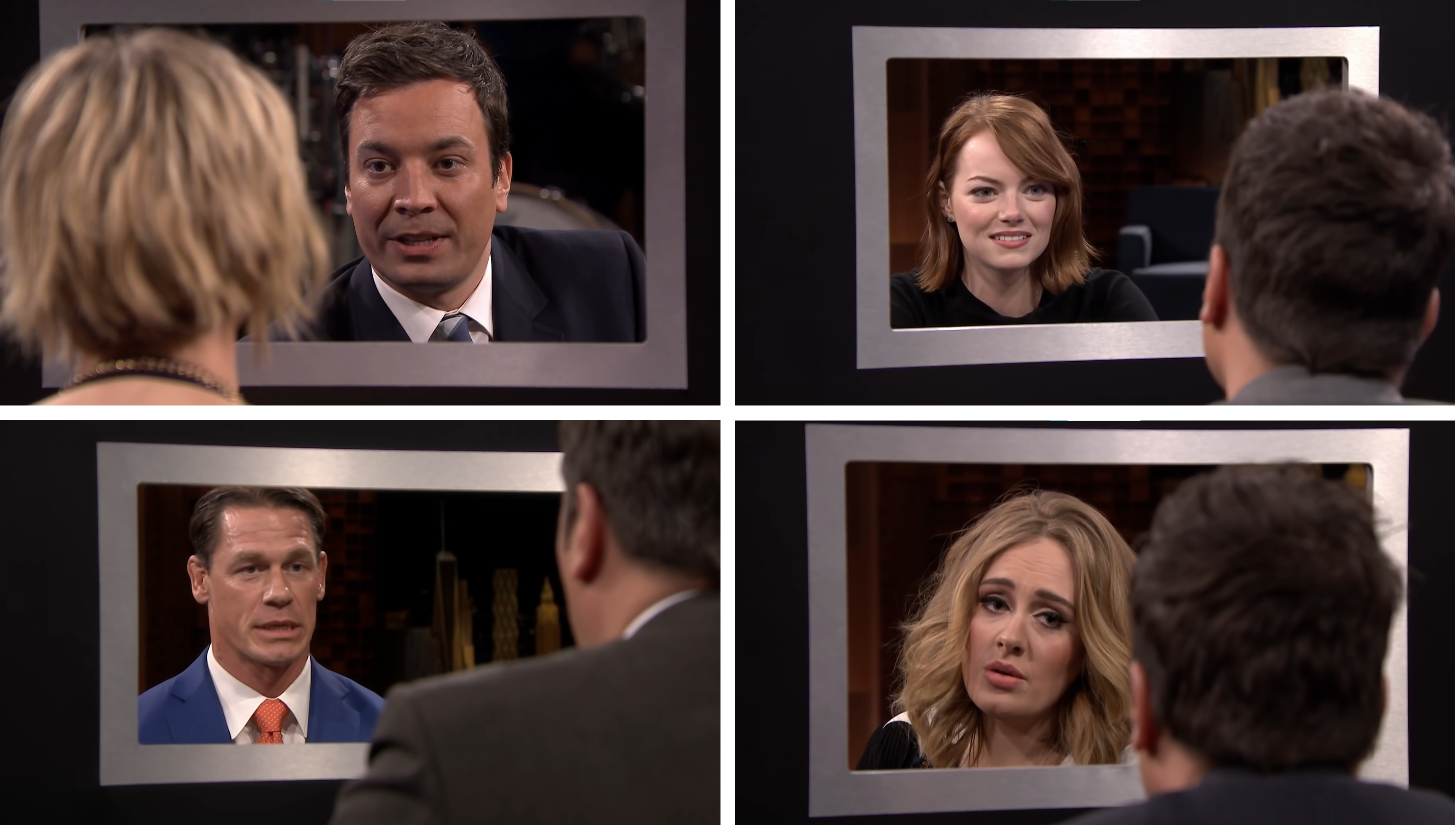}
   \caption{Sample frames from the Box of Lies dataset (BoL) \cite{Soldner2019Box} which contains videos of low-stakes deception from a television show. The top two frames are lies and the bottom two are truths.}
   \label{fig_bol}
\end{figure}

The BoL dataset \cite{Soldner2019Box} features videos from participants playing the ``Box of Lies'' game in the ``The Tonight Show Starring Jimmy Fallon \textregistered'' television show. Played between a celebrity guest and the show's host, the game consists of multiple rounds in which players describe (deceptively or truthfully) an object which is hidden from the other player, with the listener having to guess whether the description was a truth or a lie. Sample frames are presented in \cref{fig_bol}. 

The BoL dataset provides transcriptions, annotations on gestures and facial displays, as well as labels for small segments (called utterances) within each round played, with the videos being available on YouTube in an official playlist of the television show. When the Box of Lies paper \cite{Soldner2019Box} was published, 68 rounds were available, though currently this number has gone up to 93 rounds. Using these 68 rounds, the original Box of Lies paper achieves an accuracy of 65\% with a Random Forest classifier on multimodal data, while Zhang \etal \cite{Zhang2020Multimodal}, also using Random Forest, achieves 73\% accuracy for multimodal classification and 67\% for classification solely on videos.

\subsection{High-stakes datasets}
\label{subsec_high}

\begin{figure}
   \centering
   \includegraphics[width=\linewidth]{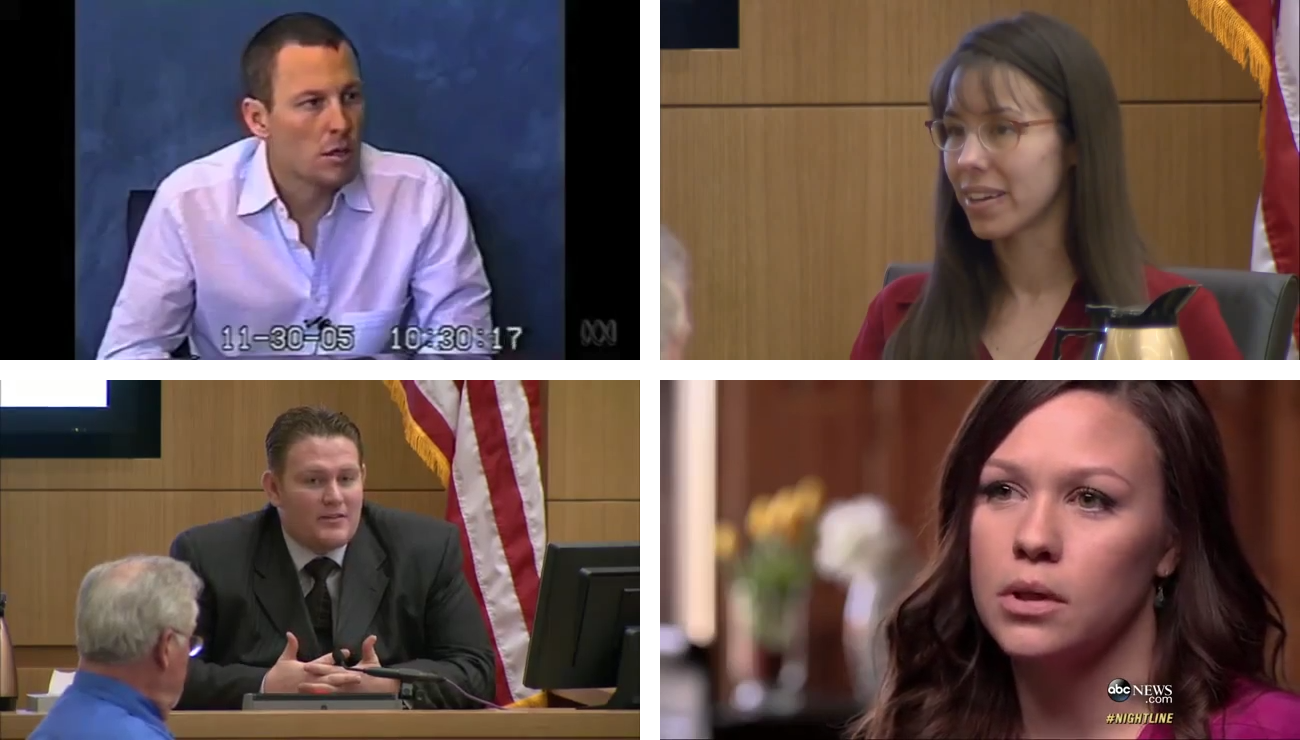}
   \caption{Sample frames from the Real-life Trial dataset (RLT) \cite{PerezRosas2015DD} which contains videos of high-stakes deception from trials. The top two frames are lies and the bottom two are truths.}
   \label{fig_rlt}
\end{figure}

To the best of our knowledge, the only publicly available video-based high-stakes DD dataset is the Real-life Trial dataset \cite{PerezRosas2015DD} (RLT), which makes it the standard dataset for papers that apply ML techniques to DD \cite{Mathur2020Introducing}. It originally consists of 121 videos from 56 unique individuals taken from trials in which truthful or deceptive testimonies were verified by the police as such, thus allowing the dataset creators to objectively label each video. It is available on one of its authors' page \cite{Mihalcea} and accompanies transcriptions for each video and annotations on non-verbal behavior such as facial displays and hand gestures. Sample frames from the dataset are presented in \cref{fig_rlt}.

Several papers have used the RLT dataset for ML experiments, which have their results and techniques summarized in \cref{tab_rlt_papers}. Among these, data augmentation strategies, classifiers, feature selection and preprocessing techniques vary greatly. With a single exception \cite{Chebbi2021Deception}, all papers that combine video, audio and transcripts achieve better results in multimodal classification than in unimodal classification from videos, showcasing the strength of combining multiple types of data for this task. But most importantly, all papers achieve results above human accuracy for DD (54\% \cite{bond2006}) and the accuracy measured for this specific set of videos by the creators of RLT in a set of experiments with human volunteers (last row of \cref{tab_rlt_papers}).

\begin{table*}[h!tb]
\centering
\begin{tabular}{|c|c|c|c|c|}
\hline
\textbf{Paper}         & \textbf{Classifier}  & \textbf{Details}                            & \textbf{\begin{tabular}[c]{@{}c@{}}Acc.\\ V\end{tabular}} & \textbf{\begin{tabular}[c]{@{}c@{}}Acc.\\ V+A+T\end{tabular}} \\ \hline \hline
\cite{Venkatesh2020Video}     & CNN + LSTM          & Vague methodology                                           & 100             & -                   \\ \hline
\cite{Karnati2022Lienet}                & CNN                 & Preprocessing with Local Binary Patterns    & 97.35           & 97.33               \\ \hline
\cite{Ding2018Face}           & CNN                 & ResNet \cite{He2016ResNet}, GANs \cite{Goodfellow2014GANs}                               & 93.61           & 97.00               \\ \hline
\cite{Krishnamurthy2018DL}    & 3DCNN               & Fusion by Hadamard product                  & 93.08           & 96.14               \\ \hline
\cite{Gogate2017DL}           & 3DCNN               & Fusion by concatenation                     & 78.57           & 96.42               \\ \hline
\cite{Ngo2018Deception}       & CNN + RNN           & Facial reconstruction                       & 72.8            & -                   \\ \hline \hline
\cite{Carissimi2018MultiView} & SVM                 & Features from AlexNet \cite{Krizhevsky2017AlexNet}, Multiview learning    & 99              & 99                  \\ \hline
\cite{Chebbi2021Deception}                 & kNN                 & Manually annotated behavioral cues          & 94              & 78                  \\ \hline
\cite{Wu2017DD}               & Logistic Regression & Features: Improved Dense Trajectory (IDT) \cite{IDT}   & 89.88*          & 92.21*              \\ \hline
\cite{Sen2022Multimodal}                    & NN          & Features: Facial displays and hand gestures & 78.53           & 84.18               \\ \hline
\cite{Avola2019Automatic}     & RBF-SVM             & Features: Action Units (AUs) \cite{ActionUnits}               & 76.84           & -                   \\ \hline
\cite{Mathur2020Introducing}  & SVM                 & Features from OpenFace \cite{OpenFace}, Affect \cite{Affect}              & 76              & 84                  \\ \hline
\cite{Karimi2018Toward}                 & LMNN                & Features from CNN + LSTM network              & 75              & 84.16               \\ \hline
\cite{Yang2020Emotion}                   & RF, SVM             & Emotion classification                      & 71.15           & 87.59               \\ \hline
\cite{PerezRosas2015DD}                  & DT                  & Features: Facial displays and hand gestures & 68.59           & 75.20               \\ \hline
\cite{Jaiswal2016Truth}       & SVM                 & Features: AUs \cite{ActionUnits}                              & 67.2            & 78.95               \\ \hline
\cite{Mambreyan2022Dataset}              & Linear SVM          & Features: IDT \cite{IDT}, Gender classifier            & 64.6            & -                   \\ \hline
\cite{Mambreyan2022Dataset}              & Linear SVM          & Features: IDT \cite{IDT}                                & 57.4            & -                   \\ \hline \hline
\cite{PerezRosas2015DD}                  & Human performance   & Average between 3 annotators                & 46.50           & 56.47               \\ \hline
\end{tabular}
\caption{Comparison of papers that use the Real-life Trial dataset \cite{PerezRosas2015DD} for Deception Detection with Machine Learning. Accuracy for classification exclusively from videos (Acc. V) is presented, as well as multimodal classification with Video, Audio and Transcripts (Acc. V+A+T). Marked accuracies (*) are actually AUC. Papers that use exclusively Deep Learning classifiers are grouped on top for convenience. All classifiers achieve better results than the measured human performance for this dataset presented on the bottom row.}
  \label{tab_rlt_papers}
\end{table*}

However, according to Belavadi \etal \cite{Belavadi2020Multimodal}, ``there is insufficient evidence that AI systems for detecting deception are likely to achieve adequate accuracy in real-world use'' and, given that the RLT dataset is small, such impressive experimental results do not provide such evidence. Moreover, Mambreyan \etal \cite{Mambreyan2022Dataset} have shown that the RLT dataset has significant gender bias which can be exploited by an ML classifier, further highlighting the need for bigger, more diverse and less biased publicly available video-based high-stakes DD datasets.

\section{Methodology}
\label{sec_methodology}

To evaluate a possible distinction between low-stakes and high-stakes deception from a Machine Learning (ML) perspective, the Box of Lies (BoL) \cite{Soldner2019Box} and Real-life Trial (RLT) \cite{PerezRosas2015DD} datasets were selected. While RLT is, to the best of our knowledge, the only publicly available video-based high-stakes Deception Detection (DD) dataset, BoL is the low-stakes dataset that has the closest number of videos to RLT (considering its current number of 93 rounds), making it a fairer comparison than other low-stakes datasets which have almost thrice the number of videos of RLT.

\subsection{Dataset preparation}
\label{subsec_dataset_prep}

Following the steps taken by other papers that work with RLT \cite{Ngo2018Deception, Ding2018Face, Wu2017DD, Mathur2020Introducing, Jaiswal2016Truth, Carissimi2018MultiView}, we remove videos from the dataset which are deemed unsuitable for classification based solely on visual data, such as those in which the speaker's face is hidden or out of focus during most of the video, or where there are multiple people in the foreground. After that, the dataset was reduced to 110 videos portraying 51 individuals. Among the remaining videos, some had noisy frames or frames showing other people instead of the speaker, so they were edited with Kdenlive \cite{Kdenlive} and these frames were removed. The full list of removed and edited videos is available in the Supplementary Material.

While the BoL dataset was originally labeled by utterance, the RLT dataset is labeled by video, which required a change in labels in BoL. Therefore, each round was taken as a single video labeled according to the veracity of the object's description and, for each video, the frames not showing the speaker were removed with Kdenlive \cite{Kdenlive}. Despite assigning a single label to whole videos (which contain truthful and deceptive segments), such labeling scheme follows the scheme from RLT (allowing for better comparison between datasets) and removes subjectivity from the process. By contrast, the original labeling by utterance categorized segments with unclear veracity as deceptive, which resulted in 82.2\% of the samples being labeled as deceptive.

After this dataset preparation step, the resulting BoL and RLT datasets were such as presented in \cref{tab_rlt_vs_bol_data}, from which it can be inferred that an ML system trained in RLT should have better results than a similar one trained in BoL, since RLT has more data while also being more diverse (more individuals) and balanced. To illustrate such balance, the individual with the highest number of videos in RLT is depicted in 21 videos (19.1\% of the videos) while for BoL, this number goes up to 34 videos (36.6\% of the videos).

\begin{table}
\centering
\begin{tabular}{c|cc}
\hline 
            & \textbf{Real-life Trial} & \textbf{Box of Lies} \\ \hline 
Videos                                   & 110             & 93          \\ \hline
Individuals                          & 51              & 33          \\ \hline
\begin{tabular}[c]{@{}c@{}}Average number of\\videos by individual\end{tabular}     & 2.2             & 2.8         \\ \hline
\begin{tabular}[c]{@{}c@{}}Standard deviation of\\videos by individual\end{tabular} & 3.5             & 5.6         \\ \hline
\begin{tabular}[c]{@{}c@{}}Average\\ video length\end{tabular}                       & 26.9 s          & 20.8 s  \\ \hline 
\end{tabular}
\caption{Comparison between the Real-life Trial (RLT) \cite{PerezRosas2015DD} and Box of Lies (BoL) \cite{Soldner2019Box} datasets after dataset preparation. Each video portrays a single individual. Not only does RLT have more videos depicting a greater number of individuals with a longer average length, but it also is more balanced in regards to the number of videos by individual (a higher standard deviation of videos by individual indicating a less balanced dataset).}
  \label{tab_rlt_vs_bol_data}
\end{table}

Also, considering the findings from Mambreyan \etal \cite{Mambreyan2022Dataset}, we consider biases present in each dataset, which are shown in \cref{tab_rlt_vs_bol_bias}. Despite not being a desirable property of the dataset, RLT has a significant gender bias, which should also make ML systems trained on it have better results due to bias exploitation.

\begin{table}
\centering
\begin{tabular}{c|cc}
\hline 
            & \textbf{Real-life Trial} & \textbf{Box of Lies} \\ \hline 
Truth ratio                                 & 51.8\%          & 46.3\%      \\ \hline
Women ratio                                 & 61.8\%          & 48.4\%      \\ \hline
\begin{tabular}[c]{@{}c@{}}Truth ratio\\ for women\end{tabular}                      & 35.3\%          & 53.3\%      \\ \hline
\begin{tabular}[c]{@{}c@{}}Truth ratio\\ for men\end{tabular}                        & 78.6\%          & 39.6\%      \\ \hline 
\end{tabular}
\caption{A comparison of biases in the Real-life Trial (RLT) \cite{PerezRosas2015DD} and Box of Lies (BoL) \cite{Soldner2019Box} datasets after dataset preparation. Even though BoL has a slight bias toward lies, RLT has a significant gender bias.}
  \label{tab_rlt_vs_bol_bias}
\end{table}

As a final preparation step, both datasets are enriched with copies of their videos flipped horizontally, doubling the number of videos in each dataset.

\subsection{Network training}
\label{subsec_network}

Focusing solely on DD from videos (as discussed in \cref{sec_datasets}), we use the Slowfast \cite{Feichtenhofer_2019_ICCV} architecture, pre-trained on the Kinetics-400 \cite{kinetics400} dataset, to perform binary classification. This architecture was chosen for its balance between classification accuracy and low computational cost, achieving better results than other Deep Learning video recognition architectures that have similar computational cost, such as Hidden TSN \cite{Zhu2019Hidden}, TSM \cite{Lin2019TSM}, TEINet \cite{Liu2020TEINet}, MSNet \cite{kwon2020motionsqueeze}, TEA \cite{Li2020TEA}, STM \cite{Jiang2019STM} and CSN \cite{Tran2019Video}.

Such criterion was used due to limited computational power and time allocation of resources for our experiments, which were performed in a shared computer with an  Intel Core i3-10100F 3.6 GHz CPU, 16 GB 2666 MHz RAM and a GTX 1650 4 GB GPU. These limitations also prompted us to use the Slowfast implementation available in the GluonCV framework \cite{gluoncv2020} for its optimizations and ease of implementation, as well as performing non-exhaustive hyperparameter search prior to k-fold testing.

For each network trained, hyperparameter search was performed independently, with the best 5 hyperparameter combinations later being used with 5-fold testing. Since non-exhaustive search was performed, the 5 best combinations were used for testing as a means of overcoming a possible deficit from the non-exhaustive search. An 80/20 split was performed for hyperparameter search, with an initial set of values being tested, and the following combinations being chosen according to the previous combinations' accuracy on the validation set. Therefore, hyperparameter values that did not perform well were assessed but soon discarded. Such strategy meant not all hyperparameter combinations were evaluated, but results on the validation set were considered satisfactory given that each combination had its results thoroughly analyzed.

Hyperparameters evaluated and their corresponding values were: 
\begin{itemize}
    \item Slowfast configuration: 4x16 and 8x8;
    \item Optimizer: SGD, Adam and RMSProp;
    \item Learning rate: ranging from $10^{-1}$ to $10^{-6}$ in exponential increments;
    \item Weight decay: $10^{-2}$, $10^{-4}$, $10^{-6}$ and no weight decay;
    \item Learning rate decay strategy: reducing learning rate by a factor of 10 every 40 epochs, every 10 epochs, or not reducing it at all;
    \item Momentum: 0.98, 0.9 and 0.5.
\end{itemize}

For most trials, 100 epochs were used, but for combinations that had not converged by 100 epochs, greater numbers were also evaluated. The batch size was limited to 1 since there was not enough memory available for greater values, a limitation which also inhibited the use of different Slowfast backbones apart from ResNet50 \cite{He2016ResNet}. The list of all hyperparameter combinations evaluated is available in the code repository for this paper.

\subsection{Experimental setup}
\label{subsec_setup}

\begin{figure*}[t]
   \centering
   \includegraphics[width=\linewidth]{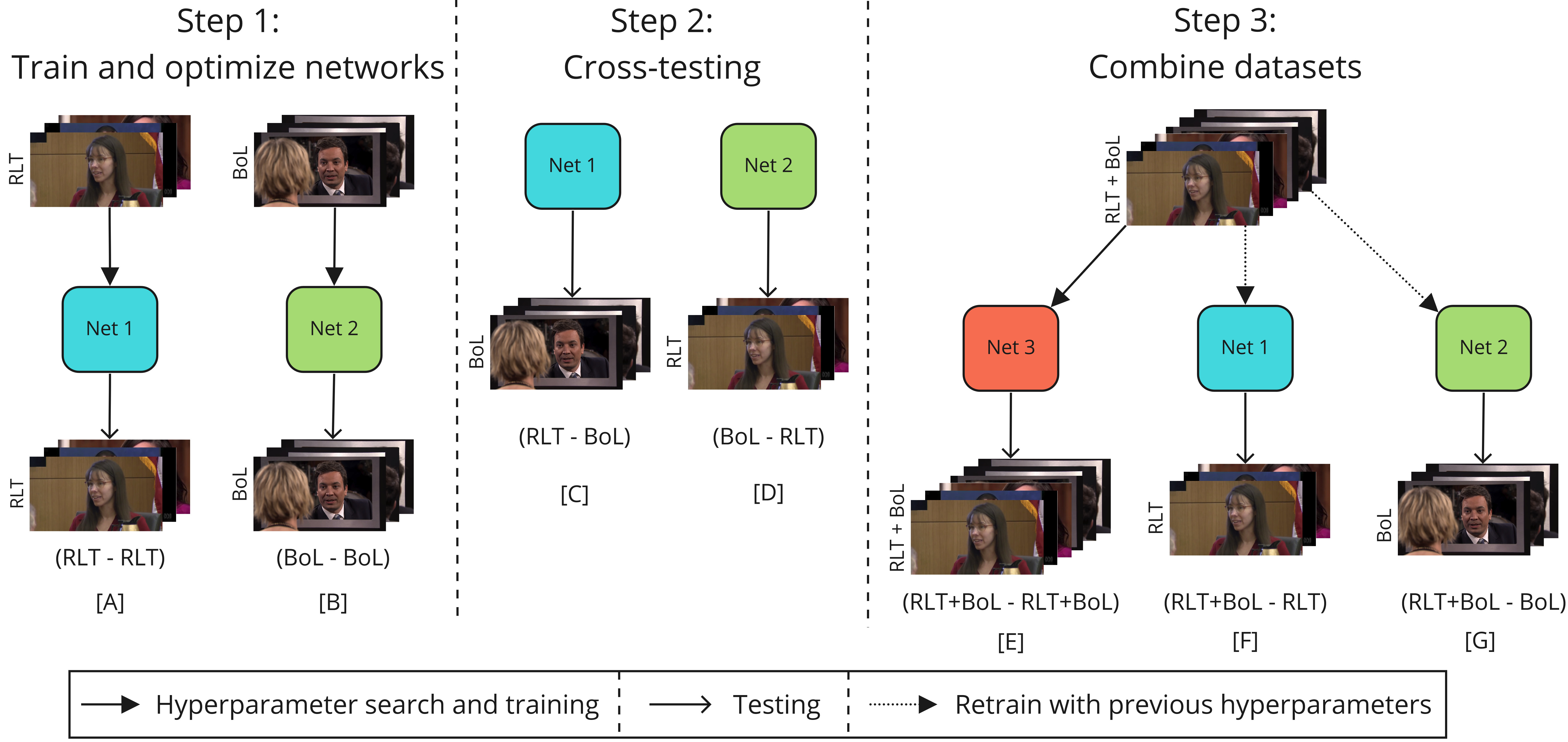}
   \caption{Experimental setup used to evaluate differences between a low-stakes deception dataset (Box of Lies - BoL \cite{Soldner2019Box}) and a high-stakes one (Real-life Trial - RLT \cite{PerezRosas2015DD}). The network used was Slowfast \cite{Feichtenhofer_2019_ICCV}, with three distinct training configurations based on the hyperparameter searches performed for the distinct training sets (RLT, BoL and RLT+BoL). Letter tags are used in each Experiment for later reference.}
   \label{fig_setup}
\end{figure*}

Our experimental setup is presented in \cref{fig_setup}. As a first step, hyperparameter search is performed considering each dataset independently. The 5 hyperparameter combinations with the highest validation accuracies are then used for 5-fold testing with their respective datasets. Even though both use the Slowfast architecture, the combination that yields the best accuracy for RLT is named Net 1 while the one which yields the best accuracy for BoL is named Net 2. Both are considered optimized for their respective datasets.

Trained on the RLT dataset, Net 1 is then used for inference on the BoL dataset while Net 2, having been trained on BoL, is used for inference on RLT (Step 2 in \cref{fig_setup}). Since both Net 1 and Net 2 were optimized for their datasets, this step allows us to evaluate whether each kind of deception dataset (high-stakes and low-stakes) can be used as training data for inference on the other kind.

Finally, both datasets are combined (Step 3 in \cref{fig_setup}) and hyperparameter search is performed for the mixed dataset, to evaluate the efficacy of combining the two kinds of deception datasets. The 5 best hyperparameter combinations are used for 5-fold testing and the best one is named Net 3.

Aiming to evaluate possible data augmentation strategies, both Nets 1 and 2 are retrained (using their previous set of hyperparameters) on the mixed dataset and tested with their original datasets.

\section{Results and discussion}
\label{sec_results}

Given that 5-fold testing was performed for all experiments except cross-testing, a total of 27 runs were conducted for testing. Hyperparameter search for Nets 1, 2 and 3 required 44, 176 and 77 runs, respectively. Therefore, a total of 324 runs were performed. If an exhaustive hyperparameter search was performed along with 5-fold (exhaustive search being done for each fold), the number of runs would be above 10000, which would not be viable given our previously described computing power limitations. Even though our approach does not guarantee optimal results, it achieves adequate accuracies with a fraction of the resources needed. The results from hyperparameter search runs are available on the code repository of this paper.

\subsection{Network optimization}
\label{subsec_hyperparameters}

The hyperparameter search for Nets 1, 2 and 3 was performed and the 5 best combinations of each were later used for 5-fold testing. The combination which had the best result was considered the one that better optimized its respective network. \cref{tab_best_hyperparams} presents the hyperparameter values for each of these combinations.

\begin{table*}
\centering
\begin{tabular}{c|ccccccc}
\hline
\textbf{Network} & \textbf{\begin{tabular}[c]{@{}c@{}}Slowfast\\Configuration\end{tabular}} & \textbf{\begin{tabular}[c]{@{}c@{}}Learning\\ Rate\end{tabular}} & \textbf{\begin{tabular}[c]{@{}c@{}}Learning Rate\\ Decay Strategy\end{tabular}} & \textbf{Optimizer} & \textbf{Momentum} & \textbf{\begin{tabular}[c]{@{}c@{}}Weight\\ Decay\end{tabular}} & \textbf{Epochs} \\ \hline
Net 1         & 4x16              & $5*10^{-4}$                                                      & Every 40 epochs                      & SGD                & 0.9               & $10^{-4}$                                                       & 100             \\
Net 2         & 8x8               & $10^{-3}$                                                        & Every 10 epochs                       & SGD                & 0.5               & $10^{-6}$                                                       & 100             \\
Net 3         & 4x16              & $10^{-5}$                                                        & No                                                                              & Adam               & -                 & $10^{-4}$                                                       & 200             \\ \hline
\end{tabular}
\caption{Best hyperparameter combinations for each of the networks trained after hyperparameter search and 5-fold testing were performed. Despite all networks using the same architecture and performing the same task, differences in training data caused significant differences in the hyperparameter combinations for each network.}
\label{tab_best_hyperparams}
\end{table*}

The results of the optimization step for each of the networks are presented in \cref{tab_training_sets}, from which the assumptions raised in \cref{subsec_dataset_prep} hold true: the RLT dataset (Exp. A) had better results than BoL (Exp. B). Both networks had results above human accuracy for DD (54\%), however both were below the results achieved by most works presented in \cref{sec_datasets}, which can be attributed to experimental limitations (such as a batch size of 1) and the fact that the Slowfast architecture was originally created for the Kinetics-400 \cite{kinetics400} dataset, which has considerably more data with 306245 videos, and often showed overfitting in our experiments.

\begin{table}
\centering
\begin{tabular}{cc|ccc}
\hline
\textbf{Net.} & \textbf{Exp.} & \textbf{Training} & \textbf{Testing} & \textbf{Acc.} \\ \hline
Net 1 & A & RLT               & RLT              & 68.64\%           \\
Net 2 & B & BoL               & BoL              & 55.92\%           \\
Net 3 & E & RLT+BoL           & RLT+BoL          & 59.12\%           \\ \hline
\end{tabular}
\caption{Comparison of results (Acc.) obtained in experiments (Exp.) for each of the networks (Net.). Hyperparameter search was conducted for each distinct training set to optimize results. While RLT obtained the best results, it is clear that BoL is a difficult dataset and it reduces overall accuracy if combined with RLT.}
  \label{tab_training_sets}
\end{table}

The combination of both datasets (Exp. E), however, had its results closer to those from BoL (Exp. B) than those from RLT (Exp. A), indicating that an increase in training data, by itself, is not enough to achieve better results. It is unclear to us whether these results occurred due to an improper combination of data with a significant semantic difference (low-stakes lies and high-stakes lies) or because of properties of these specific datasets (gender bias, category bias, imbalance of the number of videos by individual).

\subsection{Real-life Trial (Net 1) results}
\label{subsec_rlt_results}

The results from experiments done with Net 1 are presented in \cref{tab_rlt_results}, which highlights the previous considerations on the difficulty of each dataset. While the best results are seen when RLT is trained and tested on its own (Exp. A), cross-testing on BoL (Exp. C) had results below human accuracy, and adding BoL videos as an augmentation strategy (Exp. F) ends up reducing the original accuracy (Exp. A).

\begin{table}
\centering
\begin{tabular}{c|ccc}
\hline
\textbf{Exp.} & \textbf{Training} & \textbf{Testing} & \textbf{Accuracy} \\ \hline
A & RLT               & RLT              & 68.64\%           \\
C & RLT               & BoL              & 44.09\%           \\
F & RLT+BoL           & RLT              & 56.82\%           \\ \hline
\end{tabular}
\caption{Comparison of results obtained for experiments (Exp.) on Net 1. Cross-testing had significantly lower results, indicating the difficulty of the BoL dataset. Adding BoL samples to the training data lowered accuracy.}
  \label{tab_rlt_results}
\end{table}

Even though this cross-testing scenario does not hold much practical purpose since pragmatic interest lies in testing being performed with high-stakes lies and training with low-stakes lies, the evaluation of the augmentation strategy meets such interests. That being so, in our experiments, using low-stakes lies to augment a high-stakes deception dataset resulted in worse performance than working with the high-stakes dataset on its own. The results suggest that this behavior occurred due to semantic differences between these datasets (high-stakes and low-stakes deception). However, it is important to point out that the reduction in performance might have been caused by the aforementioned properties within the datasets which are unrelated to these semantic differences.

\subsection{Box of Lies (Net 2) results}
\label{subsec_bol_results}

Experiments performed with Net 2 are shown in \cref{tab_bol_results}, which presents noteworthy results. Even though training and testing in BoL reached a low accuracy (Exp. B), cross-testing with RLT (Exp. D) increased this accuracy and using RLT to augment data on BoL (Exp. G) significantly improved results.

\begin{table}
\centering
\begin{tabular}{c|ccc}
\hline
\textbf{Exp.} & \textbf{Training} & \textbf{Testing} & \textbf{Accuracy} \\ \hline
B & BoL               & BoL              & 55.92\%           \\
D & BoL               & RLT              & 58.64\%           \\
G & RLT+BoL           & BoL              & 62.87\%           \\ \hline
\end{tabular}
\caption{Comparison of results obtained for experiments (Exp.) on Net 2. Cross-testing improved results, indicating that RLT is an easier dataset than BoL. Adding RLT samples to the training data increased accuracy further.}
  \label{tab_bol_results}
\end{table}

In view of the considerations previously presented on the difficulty of each dataset, the results from Exp. G can be understood from the perspective that videos from RLT make training easier, allowing the network to better identify patterns within the testing dataset (BoL). From another point of view, it might be argued that high-stakes deception are more easily identified than low-stakes, making them suitable for data augmentation.

However, Exp. D's results are not entirely conclusive. It might be argued that, since training was performed in a difficult dataset, testing in an easier one would improve results. Despite that, such an experimental scenario perhaps holds the greatest practical interest due to the ease of acquiring low-stakes deception data and the difficulty of doing so for high-stakes data (which gets the most practical interest for DD systems). From our experiments alone, it could be said that an ML system trained with low-stakes lies could be acceptably applied to a set of high-stakes lies. However, due to our experimental limitations and limited data, such findings cannot be generalized for all DD datasets.

\section{Conclusion}
\label{sec_conclusion}

We set out to tackle the issue of data scarcity in Deception Detection (DD), aiming to shed light on the conceptual distinction between high-stakes and low-stakes lies by presenting numerical data that might support researchers on the task of building new datasets for such an important application area. Currently, datasets for DD are either built with high-stakes or low-stakes lies and applications trained in one kind theoretically can only be used for inference on lies of that same kind. Through our investigation, we evaluated this distinction by experimentally comparing two datasets: Real-life Trial (RLT) \cite{PerezRosas2015DD}, a high-stakes DD dataset, and Box of Lies (BoL) \cite{Soldner2019Box}, a low-stakes DD dataset. Different experiments were performed with the Slowfast \cite{Feichtenhofer_2019_ICCV} architecture and their results were used to analyze whether such a distinction holds true. From these experiments we found that:

\begin{enumerate}
    \item The network trained in high-stakes lies performed better than the network trained in low-stakes lies;
    \item Combining both datasets into a single one had worse results than working with the high-stakes dataset on its own;
    \item Using the network trained in high-stakes deception for inference on low-stakes deception had results below human accuracy;
    \item Using low-stakes deception as a data augmentation strategy for the high-stakes dataset did not improve results;
    \item The network trained in low-stakes lies had better accuracy classifying high-stakes deception than low-stakes;
    \item Using high-stakes lies as a data augmentation strategy for the low-stakes dataset showed a significant improvement in results.
\end{enumerate}

Given these findings, we conclude that there is a clear distinction between the RLT and BoL datasets. However, similar experiments should be performed with different datasets to assess whether these results stem from differences in these datasets' incidental properties (mainly bias and amount of data) or from a deep semantic difference in their data (low-stakes and high-stakes lies). Nonetheless, to the best of our knowledge, we have shown, for the first time, that a low-stakes DD dataset can be acceptably used to train a Machine Learning (ML) classifier created for inference on high-stakes deception data.

Finally, we highlight the need for bigger, less biased and more balanced publicly available video-based DD datasets. Despite impressive results in experimental settings, ML DD systems are not yet ready for deployment in real-life settings given the lack of sufficient data for their proper training.

{\small
\bibliographystyle{ieee_fullname}
\bibliography{paper}
}

\end{document}


\title{Supplementary Material for ``Can lies be faked? Comparing low-stakes and high-stakes deception video datasets from a Machine Learning perspective''}

\author{Mateus Karvat Camara\\
Universidade Estadual do Oeste do Paraná\\
{\tt\small mkarvat@gmail.com}
\and
Adriana Postal\\
Universidade Estadual do Oeste do Paraná\\
{\tt\small adriana.postal@unioeste.br}
\and
\hspace{1.2cm}Tomas Henrique Maul             \\
\hspace{1.2cm}University of Nottingham           \\
\hspace{1.2cm}{\tt\small tomas.maul@nottingham.edu.my}
\and
\hspace{0.8cm}Gustavo Paetzold\\
\hspace{0.8cm}Universidade Tecnológica Federal do Paraná\\
\hspace{0.8cm}{\tt\small ghpaetzold@utfpr.edu.br}
}
\maketitle

\section{Changes in the Real-life Trial dataset}

The Real-life Trial dataset \cite{PerezRosas2015DD} originally had 121 videos, 61 of which portrayed lies and 60 portrayed truths. The videos portraying lies were named as ``trial\_lie\_0XX'', with XX ranging from 01 to 61, while videos portraying truths were named as ``trial\_truth\_0XX'', with XX ranging from 01 to 60.

Based on that, the following videos were removed from the dataset for being considered unsuitable for classification based solely on videos, which requires the face of the speaker to be clearly visible:

\begin{itemize}
    \item trial\_lie\_035: shows 6 individuals;
    \item trial\_lie\_045: shows 3 individuals in the foreground;
    \item trial\_lie\_050: video from a TV show in which the individual is being questioned. A big text box makes it difficult to visualize the individual, and the back of the interrogator's head hides the speaker's face for a significant portion of the video;
    \item trial\_lie\_052: video from a TV show with many cuts, use of different angles and showing the face of the TV show's host;
    \item trial\_lie\_053: video from a YouTube channel with visual elements overlapping a video from a TV show. The video has many cuts, visual elements that can be considered as noise, as well as showing the liar for only a brief period and with the face not clearly visible;
    \item trial\_lie\_055: video shows 2 individuals in the foreground;
    \item trial\_lie\_056: video shows 2 individuals in the foreground;
    \item trial\_lie\_060: video shows 4 individuals, 2 of which in the foreground in an outdoors setting;
    \item trial\_truth\_026: video shows 4 individuals, with 2 in the foreground;
    \item trial\_truth\_029: individual's face out of focus during most of the video;
    \item trial\_truth\_041: individual's face out of focus during most of the video.
\end{itemize}

After removing these videos, the dataset amounted to 110 videos, 57 portraying truths and 53 portraying lies. Among these, a few had small parts that were considered unsuitable for classification, either for portraying other individuals instead of the speaker, either for not focusing on the speaker or for having visual noise. As such, these videos had the following edits done with Kdenlive \cite{Kdenlive}:

\begin{itemize}
    \item trial\_lie\_005: frames with visual noise cut from second 27;
    \item trial\_lie\_010: first 3 seconds cut;
    \item trial\_lie\_017: cut from second 19 to second 30;
    \item trial\_lie\_021: initial and final frames cut;
    \item trial\_lie\_022: final frames cut;
    \item trial\_lie\_023: initial frames cut;
    \item trial\_lie\_037: frames with visual noise cut from second 20;
    \item trial\_lie\_041: initial and final frames cut;
    \item trial\_lie\_042: final frames cut;
    \item trial\_lie\_043: cut from second 7 to the end of the video;
    \item trial\_lie\_044: initial and final frames cut;
    \item trial\_lie\_046: first 6 seconds cut;
    \item trial\_lie\_047: final frames cut;
    \item trial\_lie\_051: initial and final frames cut;
    \item trial\_lie\_054: cut from second 10 to the end of the video;
    \item trial\_lie\_057: initial frames cut;
    \item trial\_lie\_058: final frames cut;
    \item trial\_lie\_061: initial and final frames cut;
    \item trial\_truth\_001: final frames cut;
    \item trial\_truth\_006: initial frames cut;
    \item trial\_truth\_007: cut from second 9 to second 12 and from second 44 to second 46;
    \item trial\_truth\_008: initial and final frames cut;
    \item trial\_truth\_009: initial frames cut;
    \item trial\_truth\_042: first 9 seconds cut;
    \item trial\_truth\_051: initial frames cut.
\end{itemize}

\section{Code}

The code and the results of hyperparameter search runs are available on a GitHub \cite{github_repo}.

{\small
\bibliographystyle{ieee_fullname}
\bibliography{supplementary}
}